\documentclass[11pt,a4paper]{article}
\usepackage[hyperref]{acl2019}
\usepackage{times}
\usepackage{latexsym}
\usepackage{multicol}
\usepackage{paralist}

\usepackage{url}

\usepackage[utf8]{inputenc}

\aclfinalcopy 
\def\aclpaperid{2800} 

\usepackage{siunitx}

\usepackage{pmboxdraw}

\usepackage[disable]{todonotes} 
\newcommand{\todoin}[2][]{\todo[inline,#1]{#2}}
\newcommand{\todoinb}[2][]{\todo[inline,color=blue!40,#1]{#2}}

\def\rrr#1{{\color{red}#1}}

\newrobustcmd{\B}{\fontseries{b}\selectfont}

\newcommand{\trans}[3]{#1\textsubscript{#3} “#2”}
\newcommand{\transex}[4]{\trans{#1}{#2}{#3} (#4)}

\title{Derivational Morphological Relations in Word Embeddings}

\author{Tomáš Musil \and Jonáš Vidra \and David Mareček \\ 
  Charles University, Faculty of Mathematics and Physics \\
  Institute of Formal and Applied Linguistics \\
  Malostranské náměstí 25, 118 00 Prague, Czech Republic \\
  {\tt \{musil,vidra,marecek\}@ufal.mff.cuni.cz}}

\date{}

\begin{document}
\maketitle

\begin{abstract}

Derivation is a type of a word-formation process which creates new words
from existing ones by adding, changing or deleting affixes.
In this paper, we explore the potential of
word embeddings
to identify properties of word derivations in the morphologically rich Czech
language.
We extract derivational relations between pairs of words from DeriNet,
a Czech lexical network, which organizes almost one million
Czech lemmata into derivational trees.
For each such pair, we compute the difference
of the embeddings of the two words,
and perform unsupervised clustering of the resulting vectors.
Our results show that these clusters largely match manually annotated semantic categories
of the derivational relations (e.g.\ the relation `bake--baker' belongs to category `actor',
and a correct clustering puts it into the same cluster as `govern--governor').




\end{abstract}

\section{Introduction}

Word embeddings are a way of representing discrete words 
in a continuous space.
Embeddings are used in neural networks trained for various tasks, e.g. in neural machine translation (NMT), or can be pre-trained in various versions of language models to be used as continuous representations of words for other tasks.\todo{citace vseho moznyho}{}
One of the most popular frameworks for training word embeddings is word2vec \citep{mikolov2013efficient}.

\todo{jaky vsechny veci embeddingy zachycujou - vsechny mozny citace}

In this paper, we examine whether the word embeddings (trained on the whole words, not using any subword units or individual characters) capture derivational relations.
We do this to better understand what different neural networks represent about
words and to provide a base for further development of derivational networks.

\todoin{vyřešit opakování}

Derivation is a type of word-formation process which creates new words from
existing ones by adding, changing or deleting affixes.
For example, the word
``collide'' can be used as a base for deriving e.g.\ the words ``collider''
or ``collision''. The derived word ``collision'' can be, in turn, used
as a base for ``collisional''.

\begin{figure}
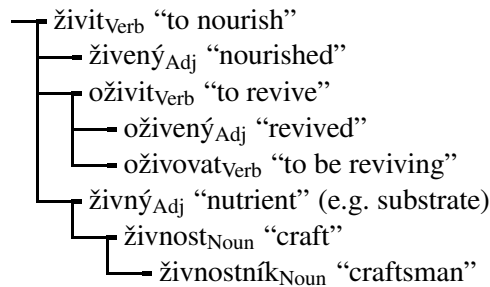

\verb|─┮| \trans{živit}{to nourish}{Verb} \\
\verb| ├─╼| \trans{živený}{nourished}{Adj} \\
\verb| ├─┮| \trans{oživit}{to revive}{Verb} \\
\verb| │ ├─╼| \trans{oživený}{revived}{Adj} \\
\verb| │ └─╼| \trans{oživovat}{to be reviving}{Verb} \\
\verb| └─┮| \transex{živný}{nutrient}{Adj}{e.g.\ substrate} \\
\verb|   └─┮| \trans{živnost}{craft}{Noun} \\
\verb|     └─╼| \trans{živnostník}{craftsman}{Noun} 

\caption{An excerpt from a derivational family rooted in the word “živit”
(to nourish, to feed).
Note that the word “oživený” (revived, rejuvenated), which can be derived from either
“oživit” (to revive) or “živený” (nourished, fed), is arbitrarily connected only
to the former, in order to simplify the derivational family to a rooted tree.}
\label{fig:example-tree}
\end{figure}

Words derived from
a single root create derivational families, which can be approximated by
directed acyclic graphs or (with some loss of information) trees; see
Figure~\ref{fig:example-tree} for an example.


Derivational relations have two sides: form-based and semantic.
For a pair of words to be considered derivationally related, the two words
must be related both by their phonological or orthographical forms
and by their meaning.


\todoin{In this paper we explore the relation between the formal and semantical
side of derivations on Czech data.}

\section{Related work}

We have not found any prior work aimed specifically at derivational relations in word embeddings.

\citet{cotterell2018joint} present a  model  of  the  semantics and structure  of  derivationally  complex  words. Our work differs in that we are examining how are derivational relations represented in preexisting applications.

\citet{gladkova2016analogy} detect morphological and semantic relations (including some derivational relations) with word embeddings. Their approach is analogy-based and they conclude that their “experiments show that derivational and lexicographic relations  remain  a  major  challenge”.

\citet{gabor2017exploring} explore vector spaces for semantic relations, using unsupervised clustering. They evaluate the clustering on 9 semantic relation classes. Our approach is similar, but we focus on derivational relations.

\citet{soricut2015unsupervised} use word embeddings to induce morphological segmentation in an unsupervised manner. Some of the relations between words that this approach implicitly uses are derivational. 

\section{Data}

In this section, we describe the network of derivational relations and the corpora used in our experiments.

\subsection{DeriNet}

There are several large networks of derivational relations available for use
in research, e.g.\ CELEX for Dutch, English and German \citep{baayen1995},
Démonette for French \citep{hathout2014},
DeriNet for Czech \citep{sevcikova2014} or 
DErivBase for German \citep{zeller2014}. A more complete listing was published
by \citet{kyjanek2018}.

For our research, we chose to use the DeriNet-1.6 network mainly due to its large
size – with over a million lemmata (citation forms),
it is over three times larger than the second
largest resource listed by \citet{kyjanek2018}, DErivBase with 280,336 lemmata.
Also, the authors are native speakers of Czech, which was necessary
for the annotation of derivation classes (see Section~\ref{annotation-of-classes} below).
Large corpora are available for Czech \citep{czeng16:2016, hnatkova2014syn}, which we need for training the word embeddings.

DeriNet is a network which approximates derivational families using trees
– the lemmata it contains are annotated with a single derivational parent
or nothing in case the word is either not derived or a parent has not been
assigned yet. It contains 1,025,095 lemmata connected by 803,404
relations.

There is a fine line between derivation and inflection and in general,
these processes are hard to separate from each other \citep[see e.g.][]{tenhacken2014}.
Both change base words using affixes, but they differ in the type of the outcome:
derivation creates new words, inflection only creates forms of the base word.
DeriNet differentiates derivation from inflection the same way
the Czech morphological tool MorphoDiTa \citep{strakova2014} does – it
considers the processes handled by the MorphoDiTa tool to be inflectional
and other affixations derivational.
This is in line with the Czech linguistic tradition \citep{dokulil1986},
except perhaps for the handling of the two main borderline cases,
whose categorization varies: negation (considered inflectional by us)
and verbal aspect changes (considered derivational).

\subsection{Word Embeddings}

In our experiments, we compare the word embeddings obtained by the standard word2vec skip-gram model~\citep{mikolov2013efficient} with word embeddings learned when training 
three different neural machine translation (NMT) models~\citep{sutskever14,bahdanau15,vaswani17}. The size of word embeddings is 512 for all the models. 

NMT models are trained between English and Czech in both directions. We use the CzEng 1.6 parallel corpus~\citep{czeng16:2016}, section \emph{c-fiction} (78 million tokens) and the Neural Monkey toolkit~\citep{vopice}\footnote{\url{https://github.com/ufal/neuralmonkey}} for training the models.
We experiment with three architectures:
\begin{itemize}
    \item \emph{RNN}: a simple recurrent neural (RNN) architecture~\citep{sutskever14} without attention mechanism, LSTM size 1,024
    \item \emph{RNN+a}: RNN architecture with attention mechanism~\citep{bahdanau15}, and
    \item \emph{Transf.}: the Transformer (big) architecture \citep{vaswani17} with 6 layers, hidden size 4,096 and 16 attention heads.
\end{itemize}
Unlike the standard setting in which embeddings of the source and the target words are shared in a common vector space, we use two separated dictionaries (each containing 25,000 word forms). We also do not use any kind of sub-word units. By this setting, we assure that the word vectors are not influenced by any other words that do not belong to the examined language.
We extract the encoder word-embeddings from Czech-English NMT model and the decoder word-embeddings from the English-Czech model.

The word2vec system is trained on the Czech National Corpus~\citep{hnatkova2014syn},
version \emph{syn 4},
which has 4.6 billion tokens.
It is a common practice \citep{mikolov2013efficient}\todo{citovat víc}{} to normalize the resulting vectors, so that the length of each vector is equal to~1. We report results for both normalized
and non-normalized
vectors.
In order to compare word2vec model with NMT models, we also train word2vec on the Czech part of the data used for training the NMT models.






All the word embeddings are trained on the word forms.
\todo{tedy nikoli na lemmas? aby se to mohlo porovnat s tim prekladem}{}
To assign an embedding to the
lemma from DeriNet, we simply select the embedding of the word form which is the same as the given lemma.

\section{Annotation of Derivational Relations}
\label{annotation-of-classes}

The derivational relations in DeriNet are not labelled in any way. In this section, we describe a simple method of automatic division of relations into \emph{derivation types} according to changes in prefixes and suffixes and then manual merging of these types into \emph{derivation classes}.

When assigning a derivation type to a relation, we first identify the longest common substring of the two related words. For instance, for the relation ``padat $\to$ padnout'', the longest common substring is ``pad''. Then, we identify prefixes and suffixes using the `+' sign for addition and `-' sign for deletion.
A sign after the string indicates a prefix and a sign before the string indicates a suffix.
Our example ``padat $\to$ padnout'' would therefore belong to the derivation type ``-at +nout'', which means deleting the suffix ``at'' and adding the suffix ``nout''. Derivation type ``na+'' means to add the prefix ``na'', etc.

\begin{table*}
\centering
\begin{tabular}{ccp{5.7cm}}
  POS & class & syntactic change \\
  \hline
A$\to$D & adjective$\to$adverb & -ý +y, -í +ě, -ý +ě, -ý +e \\
A$\to$N & designation & -ý +ec, -ý +ka \\
A$\to$N & feature & -í +ost, -ý +ost \\
A$\to$N & subject & -ký +tví \\
N$\to$A & pertaining to & +ový, -a +ový, +ní, -a +ní, -ce +ční, +ný, +ský, \mbox{-e +cký}, -ka +cký \\
N$\to$A & possessive & +ův, -a +in, -o +ův, -ek +kův, -a +ův \\
N$\to$N & diminutization & +ek, -k +ček \\
N$\to$N & instrument / scientist & -ie \\
N$\to$N & man$\to$woman & -a +ová, +ka, +ová, +vá, -ý +á, \mbox{-ík +ice} \\
N$\to$N & man$\to$woman / diminutization & -a +ka \\
N$\to$N/A$\to$A & super & super+ \\
N$\to$V & noun$\to$verb & +ovat \\
V$\to$A & ability & +elný \\
V$\to$A & acting & -it +ící, -ovat +ující, -t +jící \\
V$\to$A & general property & -t +vý \\
V$\to$A & patient & -t +ný, -it +ený, -it +ěný, -nout +lý, -t~+lý, -out +utý \\
V$\to$A & purpose & -t +cí \\
V$\to$N & actor & +el, -t +č \\
V$\to$N & nominalization & -t +ní, -at +ání, -it +ení, -it +ění, -out +utí, -ovat +ace \\
V$\to$V & imperfectivization & -at +ávat, -it +ovat \\
V$\to$V & perfectivization & -at +nout, do+, na+, o+, od+, po+, pro+, pře+, při+, roz+, u+, vy+, z+, za+ \\
\end{tabular}
\caption{Classes of Czech derivations.}
\label{tab:classes}
\end{table*}

When applied on the DeriNet relations, we identified 5,371 derivation types in total.
We selected only 71 most frequent types (only those that have at least 250 instances in DeriNet).\footnote{We count only such relations, for which both the lemmata occur at least 5 times in the Czech National Corpus.} After that, two annotators\footnote{The annotators are both native speakers of Czech and they worked together in one shared document.} manually merged the 71 derivation types into 21 classes.
The classes of derivations are listed in Table~\ref{tab:classes}.
The class \emph{super+} contains derivations from nouns to nouns and from adjectives to adjectives. 
Except for insignificant noise in the data, each of the rest of the classes contain only derivations for one POS pair.  

The classes were designed in a way
to separate different meanings of derivations where possible,
and keep different types with the same meaning together (e.g. `+ová' and `-a +ová', which derive feminine surnames).

\section{Unsupervised Clustering}

We want to know whether and how the derivational relations are captured in the embedding space.
We hypothetize that
in that case
\emph{the differences between embedding vectors} for the words in a derivational relation would cluster according to the classes we defined.

We perform unsupervised clustering of such differences using three algorithms:
\todoin{proc nefunguje cosine distance?}
\begin{compactitem}
    \item \textbf{kmeans}: K-means algorithm~\cite{macqueen1967},\footnote{We used standard Euclidean distance. The cosine distance does not work at all.}
    \item \textbf{agg}: Hierarchical agglomerative clustering using Euclidean
    distance and Ward's linkage criterion~\cite{ward1963},\footnote{We experiment also with other linking criteria, however, they performed much worse compared to the Ward's criterion.}
    \item \textbf{agg (cos)}: The same hierarchical agglomerative clustering, but using cosine distance instead of Euclidean.
\end{compactitem}

For each word pair $W_1$ and $W_2$, where $W_1$ is the derivational parent of $W_2$ and their embeddings $v_1$ and $v_2$, the clustering algorithm only gets the differience vector $d = v_2 - v_1$. The information about the word forms and their derivation type is only used in evaluation. 

We evaluate the clustering quality by homogeneity (H), completeness (C) and V-measure (V) \citep{rosenberg2007v}. These are entropy based methods, which can be compared across any number of clusters.
Homogeneity is a measure of the ratio of instances of a single class pertaining to a single cluster. Completeness measures the ratio of the member of a given class that is assigned to the same cluster.
\todoin{není na to jednoduchá rovnice jako definice nebo citace někoho, kdo to takto používá (třeba Gábor)?}
V-measure is computed as the harmonic mean of homogeneity and completeness scores.

Following \citet{gabor2017exploring}, we also report the accuracy (A) that would be achieved by the clustering if we assigned every cluster to the class that is most frequent in this cluster and then used the clustering as a classifier. The number of classes (cls) shows how many classes were assigned to at least one of the clusters.

\section{Results}

\begin{table}[t]
\sisetup{detect-weight,mode=text}
\centering
\begin{tabular}{lS[zero-decimal-to-integer, table-format=2]S[table-format=2.2, round-mode=places, round-precision=2]S[table-format=2.2, round-mode=places, round-precision=2]S[table-format=2.2, round-mode=places, round-precision=2]S[table-format=2.2, round-mode=places, round-precision=2]}
{Method} & {cls} & {H} & {C} & {V} & {A} \\
\hline
\multicolumn{6}{l}{\emph{normalized:}} \\
\hline
kmeans & 9.0 & 67.76526010539881 & 56.44333442653758 & \B 61.588286270268945 & \B 77.0 \\
agg & 10.0 & 62.29883614914483 & 52.87883267354505 & 57.203618829247525 & 72.81 \\
agg (cos) & 8.0 & 38.90320966242148 & 63.057527831385315 & 48.11931114498593 & 47.48 \\
\hline
\multicolumn{6}{l}{\emph{not normalized:}} \\
\hline
agg (cos) & 8.0 & 37.93307848472167 & 64.96679080276759 & 47.89880476012425 & 46.38 \\
agg & 9.0 & 41.18951819988613 & 39.91841360640463 & 40.544005675757084 & 50.09 \\
kmeans & 7.0 & 39.92164564333366 & 37.38022083764969 & 38.60915649992271 & 46.92 \\
\end{tabular}
\caption{Comparison of different clustering methods on differences of normalized and non-normalized word-vectors trained on Czech National Corpus and clustering into 21 clusters. The results are ordered according to V-measure.}
\label{tab:cnk-norm-notnorm}
\end{table}


The results on the vectors trained on Czech National Corpus and comparison of normalized and non-normalized versions are summarized in Table~\ref{tab:cnk-norm-notnorm}.
We can see that the normalization helps both clustering methods significantly.
\todoin{The effect of normalization is small when using the cosine distance. Co to znamena? Nemela by byt po normalizaci spis kosinova podobna te euklidovske?} The best method, i.e. the K-means used on the normalized word vectors is used in the next experiments.

\begin{table}[t]
\centering
\sisetup{detect-weight,mode=text}
\begin{tabular}{lS[zero-decimal-to-integer, table-format=2]S[table-format=2.1, round-mode=places, round-precision=2]S[table-format=2.1, round-mode=places, round-precision=2]S[table-format=2.1, round-mode=places, round-precision=2]S[table-format=2.1, round-mode=places, round-precision=2]}
{model} & {clust.} & {H} & {C} & {V} & {A} \\
\hline
baseline & 15.0 & 3.7885846417514957 & 2.69960002385002 & 3.152703606529995 & 30.824544582933846 \\
word2vec & 15.0 & 75.98122894897473 & 57.81969513560862 & \B 65.65985219211953 & 83.05848513902205 \\
baseline & 20.0 & 5.121443196628589 & 3.299870732147336 & 4.013648143829394 & 31.323106423777563 \\
word2vec & 20.0 & 77.00269422100718 & 54.03503755337274 & 63.497566854881654 & \B 84.2569511025887 \\
baseline & 21.0 & 5.307073913852612 & 3.367668235616616 & 4.120571069524643 & 30.87248322147651 \\
word2vec & 21.0 & 77.49727461546804 & 53.17424800083482 & 63.06685446320815 & 84.12272291466921 \\
baseline & 22.0 & 5.48592579037642 & 3.4285938913144216 & 4.219859313900697 & 30.97794822627037 \\
word2vec & 22.0 & 77.07211854378181 & 52.15046080181632 & 62.19646997952701 & 83.9693192713327 \\
baseline & 25.0 & 6.134339714388764 & 3.681503174374864 & 4.6014559722790995 & 31.409395973154364 \\
word2vec & 25.0 & 80.20058589262055 & 53.10565270724457 & 63.89491984453965 & 87.37296260786194 \\
\end{tabular}
\caption{Effect of number of clusters with K-means (averaged over 10 runs).
\todoin{Nebylo by tohle lepsi v grafu? T: mozna bylo, ale v tom pripade by to asi chtelo vybrat jednu hodnotu, kterou budeme plotovat;}
}
\label{tab:clust}
\end{table}

In Table~\ref{tab:clust}, we examine the effect of the number of clusters on the clustering quality.
We compare our models to the baseline,
in which each derivation pair is assigned to a random cluster.
The table shows that regardless of the number of clusters, the clustering on the word2vec embeddings performs better than the random baseline.
It shows that as we allow the K-means algorithm to form more clusters, the homogeneity increases and the completeness decreases. The V-measure is highest with the lowest number of clusters. This may be because the clusters are of uneven size. The accuracy on the word2vec model embeddings is highest around the number of clusters that corresponds to the number of classes in the data.  

\begin{table}[t]
\centering
\sisetup{detect-weight,mode=text}
\begin{tabular}{lS[zero-decimal-to-integer, table-format=1.1]S[table-format=2.1, round-mode=places, round-precision=2]S[table-format=2.1, round-mode=places, round-precision=2]S[table-format=2.1, round-mode=places, round-precision=2]S[table-format=2.1, round-mode=places, round-precision=2]}
{model} & {cls} & {H} & {C} & {V} & {A} \\
\hline
word2vec & 7.9 & 77.53252820063317 & 53.70464355637826 & \B 63.44816851396722 & \B 84.18024928092042 \\
RNN dec. & 6.8 & 73.0900378437845 & 52.195508167888285 & 60.890011173241 & 83.70086289549377 \\
RNN+a enc. & 6.4 & 59.443159467697924 & 44.92407753705828 & 51.14242570179473 & 76.09779482262704 \\
Transf. enc. & 6.4 & 60.30065908984187 & 44.235085527062196 & 51.02167426504583 & 78.2933844678811 \\
RNN+a dec. & 6.8 & 60.93724414656695 & 40.25416553434311 & 48.479806437089586 & 76.40460210930009 \\
RNN enc. & 6.4 & 51.90357249276045 & 45.12644640116477 & 48.25494904388797 & 70.48897411313519 \\
Transf. dec. & 5.5 & 44.20694981804943 & 30.562017766880622 & 36.135661669520125 & 63.41323106423778 \\
baseline & 2.8 & 5.371408837812099 & 3.407924621692009 & 4.170100755467194 & 31.150527325023965 \\
POS baseline & 8 & 52.63 & 100.00 & \B 68.97 & 45.83\\
\end{tabular}
\caption{Results on vectors learned by the NMT models compared to word2vec. K-means clustering with 21 clusters. The results are averaged over 10 independent runs.}
\label{tab:NMT}
\end{table}

Table~\ref{tab:NMT} presents the results of clustering the differences of embedding vectors from NMT models. The \emph{cls} columns shows how many different classes are assigned. Because some classes are more frequent than others, they may form the majority in multiple clusters. This is why random baseline assignes less than 3 different classes on average. We see that word2vec (trained on the Czech side of the parallel corpus) captures more information about derivations than NMT models. RNN models store more information in the embeddings if they do not utilize the attention mechanism. Even less information is stored in the embeddings by the Transformer architecture. This is probably because while in attention-less model the embedding is the only set of parameters directly associated with the given word, in the attention model the information can be split between embeddings and the attention weights. The transformer architecture with residual connections has even more parameters associated with a given word.  Decoder in general stores more information about relation between words in the embeddings than encoder, presumably because it partially supplies the role of a language model.

We also evaluated clustering by POS tags (\emph{POS baseline} in Table~\ref{tab:NMT}), where we created 8 clusters based on the POS tags of the parent and child words in the derivational relation. This clustering has a high V-measure, because its completeness is 100\,\% (the \emph{super+} class is not present in the NMT data and for all the other classes it holds that each member of a class has the same parent-child POS tags pair). But it has lower accuracy than all the other models (except for the random baseline), showing that the unsupervised clustering does more than just clustering by POS.

\begin{table*}
\centering
\begin{tabular}{lS[table-format=2.2, round-mode=places, round-precision=2]S[table-format=2.2, round-mode=places, round-precision=2]}
{Derivation class} & {precision} & {recall} \\
\hline
ability & 70.71369975389663 & 68.96 \\
acting & 47.98512089274644 & 61.92000000000001 \\
actor & 63.02521008403362 & 60.0 \\
adjective$\to$adverb & 87.71929824561404 & 28.0 \\
designation & 50.0 & 66.32 \\
diminutization & 24.09090909090909 & 42.4 \\
feature & 81.2870448772227 & 76.8 \\
general property & 63.21506154960174 & 69.84 \\
imperfectivization & 96.99530516431925 & 82.64 \\
instrument / scientist & 97.45856353591161 & 70.56 \\
man$\to$woman & 98.00747198007473 & 62.96 \\
man$\to$woman / diminutization & 34.87962419260129 & 47.52 \\
nominalization & 65.38821328344247 & 55.92000000000001 \\
noun$\to$verb & 96.7065868263473 & 77.52 \\
patient & 47.20000000000001 & 51.92000000000001 \\
perfectivization & 92.84916201117318 & 66.48 \\
pertaining to & 36.845039018952065 & 52.879999999999995 \\
possessive & 65.28704939919892 & 78.24 \\
purpose & 52.330226364846865 & 31.439999999999998 \\
subject & 69.54515491100858 & 84.4 \\
super & 31.97674418604651 & 26.4 \\
\end{tabular}
\caption{Precision and recall for the derivation classes. We sampled 250 examples for each class from the data and clustered them with K-means on word2vec embeddings trained on the ČNK. Results presented here are averaged over 5 runs.}
\label{tab:clsres}
\end{table*}

The data naturally contains classes with significant differences in size. To prevent the small classes from being underrepresented, we also evaluated the clustering on a dataset, where the same number of derivation pairs was sampled from each class.
Results for the experiment with classes of the same size are listed in Table~\ref{tab:clsres}.
The results show that the classification does not rely only on changes of part-of-speech. Both \emph{imperfectization} and \emph{perfectivization} classses are classified well (97\,\% precision, 83\,\% recall and 93\,\% precision, 66\,\% recall respectively), even though they are both derivation from verbs to verbs.
The only classes that have both precision and recall under 50\,\% are those being confused with \emph{diminutization}: \emph{man $\to$ woman} shares one common derivation type with \emph{diminutization}, and the class \emph{super}, which contains only the prefix “super” and is therefore opposite to \emph{diminutization}, sharing the same semantic axis.

\section{Conclusion}

Our results show that word-level word embeddings capture information
about semantic classes of derivational relations between words,
despite not having any information about the orthography or morphological makeup of the words,
and therefore not knowing about the formal relation between the words.

It is possible to cluster differences between embeddings in derivational relations,
and the assigned clusters correspond to the semantic classes of the relations.
The word2vec embeddings generally result in a better clustering
than embeddings from the NMT models, and embeddings from the decoder of a plain RNN model
perform better than those from NMT models with attention.
All these methods outperform a random-assignment clustering baseline and POS clustering baseline.


\todoin{future work: víc jazyků}

\todoinb{Chceme tohle zmiňovat? Když tak to smažte.\\~\\
A possible extension of this work would compare the word-level word embeddings
with character-level or subword embeddings\rrr{explain?} which could capture
the meaning of individual morphemes. Our preliminary research (not reported
here) suggests that FastText embeddings \cite{grave2018fasttext}, which utilize subword units,
outperform the word-level embeddings.}

\section*{Acknowledgments}

This work has been supported by the grant 18-02196S of the Czech
Science Foundation.
%
This study was supported by the Charles University Grant Agency (project No.\ 1176219).
This research was partially supported by SVV
project number 260 453.
This work has been using language resources and tools developed,
stored and distributed by the LINDAT/\discretionary{}{}{}CLARIN project of the Ministry of Education, Youth and Sports of the Czech Republic (project
LM2015071).

\bibliography{derivace}
\bibliographystyle{acl_natbib}

\end{document}